\documentclass[reviewcopy]{elsart}
\usepackage{pst-all}
\usepackage{amssymb}
\usepackage{amsmath}
\usepackage{epsfig}

\newcommand{\pder}[2]{\frac{\partial #1}{\partial #2}}
\newcommand{\tsp}[1]{{\vphantom{#1}}^{t}{#1}}
\newcommand{\gdt}[1]{\vec{\nabla #1}}

\renewcommand{\vec}[1]{\mathbf{#1}}

\begin{document}
  \begin{frontmatter}
    \title{Deformable Model with a Complexity Independent from Image Resolution}
    
    \author{J.-O. Lachaud \and B. Taton}\footnote{Email address:
      \{\texttt{lachaud|taton}\}\@\texttt{labri.fr}} %

    \address{%
      Laboratoire Bordelais de Recherche en Informatique \\
      351, cours de la Lib\'eration - 33400 TALENCE - FRANCE\\
      Phone: (+33) 5 40.00.69.00 \quad Fax: (+33) 5 40.00.66.69
    }

    \newpage
    \begin{abstract}
      We present a parametric deformable model which recovers image
      components with a complexity independent from the resolution of
      input images. The proposed model also automatically changes its
      topology and remains fully compatible with the general framework
      of deformable models. More precisely, the image space is
      equipped with a metric that expands salient image details
      according to their strength and their curvature. 
      During the whole evolution of the model, the sampling of the
      contour is kept regular with respect to this metric. By this
      way, the vertex density is reduced along most parts of the curve
      while a high quality of shape representation is preserved. The
      complexity of the deformable model is thus improved and is no
      longer influenced by feature-preserving changes in the
      resolution of input images. Building the metric requires a prior
      estimation of contour curvature. It is obtained using a robust
      estimator which investigates the local variations in the
      orientation of image gradient. Experimental results on both
      computer generated and biomedical images are presented to
      illustrate the advantages of our approach.
    \end{abstract}

    \begin{keyword}
      deformable model, topology adaptation, resolution adaptation,
      curvature estimation, segmentation/reconstruction.
    \end{keyword}
  \end{frontmatter}

  \newpage

  \section{Introduction}
  \label{sect:introduction}
  In the field of image analysis, recovering image components is a
  difficult task. This turns out to be even more challenging when
  objects exhibit large variations of their shape and topology.
  Deformable models that are able to handle that kind of situations
  can use only little {\em a priori} knowledge concerning image
  components. This generally implies prohibitive computational costs
  (see Table~\ref{table:complexities}).

  In the framework of parametric deformable models, most authors
  \cite{Delingette00,McInerney95,McInerney97} propose to investigate
  the intersections of the deformable model with a grid that covers
  the image space. Special configurations of these intersections
  characterize the self collisions of the mesh. Once
  self-instersections have been detected, local reconfigurations are
  performed to adapt the topology of the model according to its
  geometry.  To take advantage of all image details, 
  the grid and the image should have the same resolution. An other
  method \cite{Lachaud99a} consists in constraining the lengths of the
  edges of the model between two bounds. Self-collisions are then
  detected when distances between non-neighbor vertices fall under a
  given threshold. Topological consistency is recovered using local
  operators that reconnect vertices consistently. Using all image
  details requires edges to have the same size as image pixels. The
  complexities of all these methods are thus directly determined by
  the size of input data.

  In the framework of level-set methods, boundaries of objects are
  implicitly represented as the zero level set of a function $f$
  \cite{Caselles93,Caselles95,Malladi95,Yezzi97}. Usually $f$ is
  sampled over a regular grid that has the same resolution as the
  input image. Then $f$ is iteratively updated to make its zero
  level-set approach image contours. Even with optimization methods
  which reduce computations to a narrow band around evolving
  boundaries \cite{Adalsteinsson95,Strain99}, the complexity of these
  methods is determined by the resolution of the grid and hence by the
  resolution of the input image.

  \begin{table}[tb]
    \caption{%
      Complexities of the deformable models that automatically adapt
      their topology.  The length of the deformable model is denoted
      $l$, the with of a pixel is denoted $d$.  The size of the image is
      denoted $|I|$, and $k$ denotes the width of the narrow band when
      this optimization is used. This shows that the complexities of
      these algorithms are completely determined by the resolution of
      the input image.
    \label{table:complexities} }

    \begin{center}
      \begin{tabular}{l|c}
	Model & Complexity per iteration \\
	\hline
	T-Snake\cite{McInerney97} & $O(|I|)$ \\
	\hline
	Simplex mesh\cite{Delingette00} & $O(\frac{l}{d})$ \\
	\hline 
	Distance constraints\cite{Lachaud99a}) &
	$O(\frac{l}{d} \log\left(\frac{l}{d}\right))$ \\
	\hline
	Level-set \cite{Caselles93,Malladi95} & $O(|I|)$ \\
	\hline
	Level-set with narrow band \cite{Adalsteinsson95} & $O\left(k \frac{l}{d}\right)+0\left(\frac{l}{d} \log\left(\frac{l}{d}\right) \right)$ \\
      \end{tabular}
    \end{center}

  \end{table}

  In \cite{Taton02} a method is proposed to adapt the resolution of a
  deformable model depending on its position and orientation in the
  image. The main idea is to equip the image space with a Riemannian
  metric that geometrically expands parts of the image with
  interesting features.  
  During the whole evolution of the model,
  the length of edges is kept as uniform as possible with
  this new metric. 
  As a consequence, a well chosen metric results in an accuracy of the
  segmentation process more adapted to the processed image.
 
  In this first attempt the metric had to be manually given by a user. 
  This was time consuming and the user had to learn how to provide an
  appropriate metric.
  Our contribution is to propose an automated way of building metrics
  directly from images. The accuracy of the reconstruction is
  determined by the geometry of image components.
  The metric is built from the image
  \begin{enumerate}
  \item to optimize the number of vertices on the final mesh, thus
  enhancing the shape representation,
    \label{enum-1}
  \item and to reduce both the number and the cost of the iterations required
    to reach the rest position.
    \label{enum-2}
  \end{enumerate}

  Property (\ref{enum-1}) is obtained by building the metric in such a
  way that the length of the edges of the model linearly increases
  with both the strength and the radius of curvature of the underlying
  contours.

  Property (\ref{enum-2}) is ensured by designing the metric in such a
  way that a coarse sampling of the curve is kept far away from image
  details, while it progressively refines when approaching these
  features.
  
  To build a metric which satisfies these constraints, the user is
  asked for only three parameters:
  \begin{itemize}
  \item $s_\text{ref}$: the norm of the image gradient over which a contour is
    considered as reliable,
  \item $l_\text{max}$: the maximum length of the edges of the
    deformable model (this is required to prevent edges from growing
    too much which would lead to numerical instability),
  \item $l_\text{min}$: the minimum length of an edge (typically this
    corresponds to the half width of a pixel).
  \end{itemize}
  Over a sufficient resolution of the input image, the gradient
  magnitude as well as the curvature of objects do not depend on the
  sampling rate of the input image. Consequently the computational
  complexity of the segmentation algorithm is determined only by the
  geometrical complexity of image components and no longer by the size
  of input data.

  The dynamics of the deformable model is enhanced too. In places
  without image structures the length of edges reaches its
  maximum. This results in (i) less vertices and (ii) larger
  displacements of these vertices. By this way both the cost per
  iteration and the number of iterations required to reach convergence
  are reduced. 

  We point out that these enhancements do not prevent the model from
  changing its topology and that the complexity of these topology
  changes is determined by the (reduced) number of vertices. Other
  methods, such as those presented in \cite{Delingette00} and
  \cite{McInerney97} require the model to be resampled (at least
  temporarily) on the image grid. As a result, these methods cannot
  take advantage of better sampling of the deformable curve to reduce
  their complexity.

\begin{figure}[p]
    \begin{center}
      \epsfig{width=0.98\textwidth, file=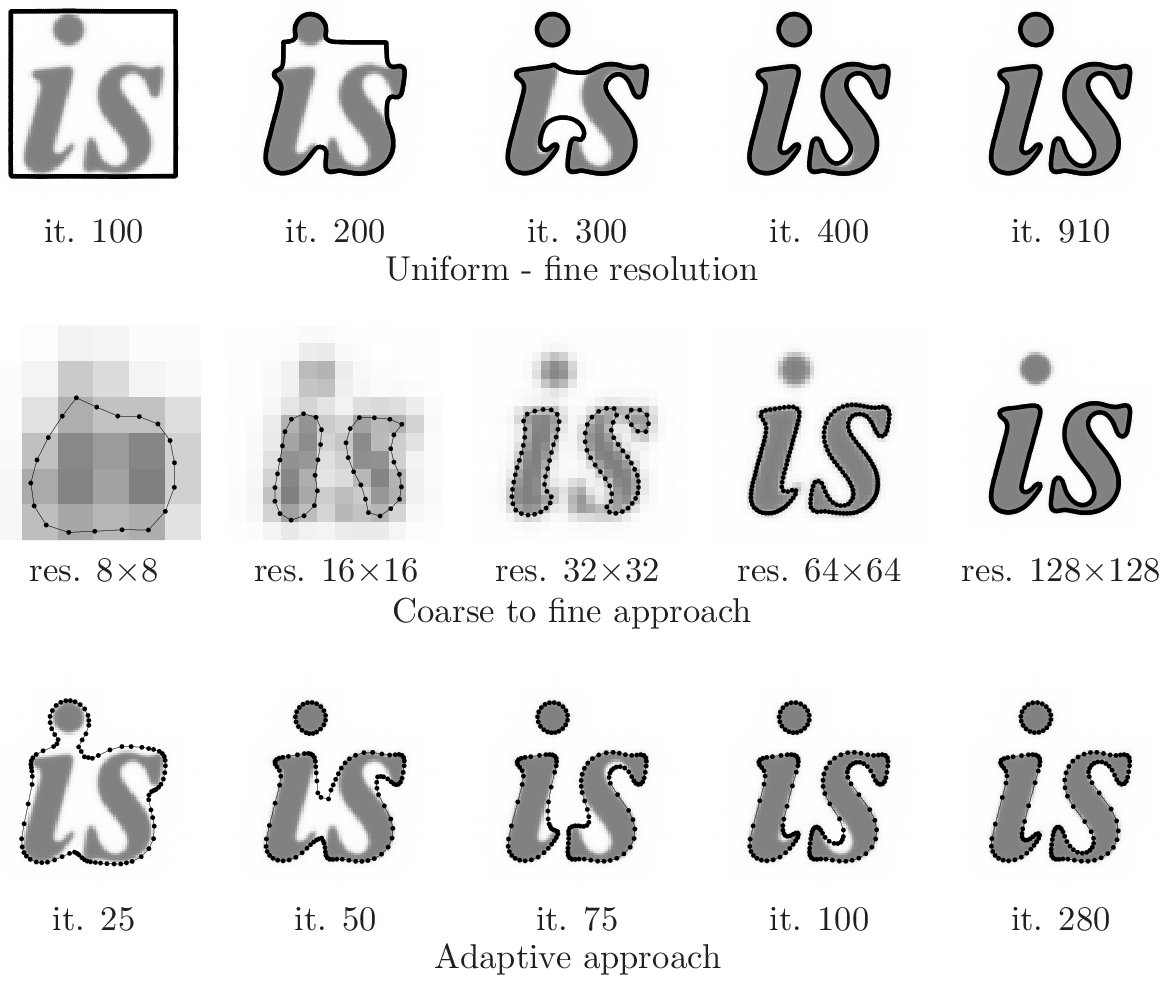}
    \end{center}

    \caption{Illustration of the proposed approach to shape
    extraction. Top row: extraction with uniform sampling of the
    model. The edge length has approximately the pixel
    width. Computation statistics: 910 iterations, 10.14s, 458
    vertices. Middle row: coarse to fine extraction. At each level,
    the edge length has approximately the pixel width. Computation
    statistics: 310+810+760+1020+510 iterations, 9.23s =
    0.08+0.40+0.87+3.10+4.78, 392 vertices. Bottom row: extraction
    with adaptive sampling of the model. The edge length varies
    between half the pixel width and 20 times the pixel
    width. Computation statistics: 280 iterations, 1.73=0.31+1.42s
    (precomputation + evolution), 150 vertices.
    \label{fig:illustration-of-proposed-approach}}
  \end{figure}

  Fig.~\ref{fig:illustration-of-proposed-approach} illustrates the
  main idea of our approach and offers a comparison with the classical
  snake approach and with a coarse-to-fine snake method. Note that the
  same parameters are used for all three experiments: force
  coefficients, initialization, convergence criterion. First, it
  appears clearly that our approach achieves the same segmentation
  quality as regular snakes with a high precision. A coarse-to-fine
  approach may fail to recover small components. Second, computation
  times are greatly improved with our approach (about 6 times
  faster). The coarse-to-fine approach is also rather slow since a lot
  of time is spent extracting details that were not present at a
  coarser level. Third, the number of vertices is optimized according
  to the geometry of the extracted shape (3 times less vertices).

  Moreover, the proposed model remains compatible with the general
  framework of deformable models: to enhance the behavior of the
  active contour, any additional force \cite{Cohen91,Xu98,Yu02} may be
  used without change. In practice, with the same sets of forces, the
  visual quality of the segmentation is better with an adaptive vertex
  density than in the uniform case. Indeed, along straight parts of
  image components, the slight irregularities that result from the
  noise in the input images are naturally rubbed out when fewer
  vertices are used to represent the shape. In the vicinity of fine
  image details an equivalent segmentation accuracy is achieved in
  both the adaptive and uniform cases.
  
  At last, the approach proposed to build the metric is almost fully
  automated. Thus, only little user interaction is required. However
  it remains easy to incorporate additional expert knowledges to
  specify which parts of image components have to be recovered
  accurately.

  This paper is structured as follows: in
  Sect.~\ref{sect:deformable-model} we describe the proposed
  deformable model and we show how changing metrics induces adaptive
  resolution. In Sect.~\ref{sect:building-metrics}, we explain how
  suitable metrics are built directly from images using a robust
  curvature estimator. Experimental evaluation of both the proposed
  model and the curvature estimator are presented in
  Sect.~\ref{sect:experiments}.

  \section{Deformable Model}
  \label{sect:deformable-model}
  \subsection{General description}
  \label{sect:general-description}
  Our proposed deformable model follows the classical energy
  formulation of active contours \cite{Kass87}: it is the
  discretization of a curve that is emdedded in the image space. Each
  of its vertices undergoes forces that regularize the shape of the
  curve, attract them toward image features and possibly tailor the
  behavior of the model \cite{Cohen91,Xu98} for more specific
  purposes. In this paper, classical parametric snakes are extended
  with the ability to (i) dynamically and automatically change their
  topology in accordance with their geometry and (ii) adapt their
  resolution to take account of the geometrical complexity of
  recovered image components. In addition, only little work is
  necessary to adapt the proposed model for three-dimensional image
  segmentation (see \cite{Lachaud03} for details). In this three
  dimensional context, the number of vertices is further reduced and
  computation times are consequently greatly improved.

  \subsection{Resolution adaptation}
  \label{sect:resolution-adaptation}
  During the evolution of the model, the vertex density along the curve
  is kept as regular as possible by 
  constraining the length of the edges of the model between two bounds
  $\delta$ and $\zeta\delta$:
  \begin{equation}
    \delta \leq L_E(u, v) \leq \zeta \delta \mbox{ ,}
    \label{eqn:constraints}
  \end{equation}
  In (\ref{eqn:constraints}) $L_E$ denotes the length of the line
  segment that joins $u$ and $v$. The parameter $\delta$ determines
  lengths of edges and hence vertex density along the curve. The
  parameter $\zeta$ determines the allowed ratio between maximum and
  minimum edge lengths.

  At every step of the evolution of the model, each edge is
  checked. If its length is found to be less than $\delta$ then it is
  contracted. In contrast, if its length exceeds the $\zeta \delta$
  threshold then the investigated edge gets split. To ensure the
  convergence of this algorithm $\zeta$ must be chosen greater than
  two. In the following $\zeta$ is set to the value $2.5$, which
  provides satisfying results in practice. The parameter $\delta$ is
  derived from the maximum edge length $l_\text{max}$ specified by the
  user as $\delta = l_\text{max} / \zeta$.
  
  Adaptive resolution is achieved by replacing the Euclidean length
  estimator $L_E$ by a position and orientation dependent length
  estimator $L_R$ in (\ref{eqn:constraints}). In places where $L_R$
  underestimates distances, estimated lengths of edges tend to fall
  under the $\delta$ threshold. As a consequence, edges tend to
  contract and the resolution of the model locally decreases. In
  contrast, the resolution of the model increases in regions where
  $L_R$ overestimates distances.

  More formally, Riemannian geometry provides us with theoretical
  tools to build such a distance estimator. In this framework, the
  length of an elementary displacement $\vec{ds}$ that starts from
  point $(x, y)$ is expressed as:
  \begin{equation}
    \|\vec{ds}\|_R^2 = \tsp{\vec{ds}} \times G(x, y) \times \vec{ds},
    \label{eqn:vector-length}
  \end{equation}
  where $G$ associates a positive-definite symmetrical matrix with
  each point of the space. The $G$ mapping is called a {\em Riemannian
    metric}. From (\ref{eqn:vector-length}) follow the definitions of
  the Riemannian length of a path as
  \begin{equation}
    L_R(\gamma) = \int_{a}^{b} \|\vec{\dot{\gamma}}(t)\|_{R} \; dt \mbox{ ,}
    \label{eqn:path-length}
  \end{equation} 
  and of the Riemannian distance between two points $u$ and $v$ as
  \begin{equation}
    d_R(u, v) = \inf_{\gamma \in {\mathcal C}} L_R(\gamma) \mbox{ ,}
    \label{eqn:riemannian-distance}
  \end{equation}
  where ${\mathcal C}$ contains all the paths that join $u$ and $v$.  It
  is thus easily seen that defining the $G$ mapping is enough to
  completely define our new length estimator $L_R$. How this mapping
  is built from images to enhance and speed up shape recovery is
  discussed in Sect.~\ref{sect:building-metrics}.

  \subsection{Topology adaptation}
  \label{sect:topology-changes}
  During the evolution of the model, care must be taken to ensure that
  its interior and exterior are always well defined: self-collisions
  are detected and the topology of the model is updated accordingly
  (see \cite{Taton02} for more details on topology adaptation).

  Since all the edges have their length lower than $\zeta \delta$, a
  vertex that crosses over an edge $(u, v)$ must approach either $u$
  or $v$ closer than $\frac{1}{2} (\zeta \delta + d_{max})$, where
  $d_{max}$ is the largest distance covered by a vertex during one
  iteration. Self-intersections are thus detected by looking for pairs
  of non-neighbor vertices $(u,v)$ for which
  \begin{equation}
    d_E(u, v) \leq \frac{1}{2} (\zeta \delta + d_{max}) \mbox{ .}
    \label{eqn:collision-detection}
  \end{equation}
  It is easily shown that this detection algorithm remains valid when
  $d_E$ is replaced with a $d_R$ distance estimator as described in
  Sect.~\ref{sect:resolution-adaptation}. With a naive implementation,
  the complexity of this method is quadratic. However, it is reduced
  to $O(n \log n)$ by storing vertex positions in an appropriate
  quadtree structure.

  Detected self-intersections are solved using local operators that
  restore a consistent topology of the mesh by properly reconnecting
  the parts of the curve involved in the collision.

  \subsection{Dynamics}
  \label{sect:dynamics}

  Theoretically, in a space equipped with a Riemannian metric, the
  position $\vec{x}$ of a vertex that undergoes a force $\vec{F}$
  follows equation
  \begin{equation}
    m \ddot{x}_k = F_k - \sum_{i,j} \Gamma_{ij}^k \dot{x}_i
    \dot {x}_j \mbox{ ,}
    \label{eqn:motion-equation}
  \end{equation}
  where the $\Gamma_{ij}^{k}$ coefficients are known as the {\em
    Christoffel's symbols} associated with the metric:
  \begin{equation}
    \Gamma_{ij}^{k} = \frac{1}{2} 
    \sum_{k} g^{kl}
    \left(
    \pder{g_{il}}{x_j} + \pder{g_{lj}}{x_i} - \pder{g_{ij}}{x_l}
    \right)
    \mbox{ .}
  \end{equation}

  However, the last term of (\ref{eqn:motion-equation}) 
  is quadratic in $\vec{\dot{x}}$ and has therefore only little
  influence when the model is evolving. Furthermore, once at rest
  position it cancels and has consequently no impact on the final
  shape. Therefore it is neglected and we get back the classical
  Newton's laws of motions. Experimentally, removing this term does
  not induce any noticeable change in the behavior of the model.

  \section{Tailoring Metrics to Images}
  \label{sect:building-metrics}

  \subsection{Geometrical interpretation}
  For any location $(x,y)$ in the image space, the metric $G(x,y)$ is
  a positive-definite symmetrical matrix. Thus, in an orthonormal (for
  the Euclidean norm) base $(\vec{v}_1,\vec{v}_2)$ of eigenvectors,
  $G(x,y)$ is diagonal with coefficients $(\mu_1, \mu_2)$. Hence, the
  length of an elementary displacement $\vec{ds} = x_1 \vec{v}_1 + x_2
  \vec{v}_2$ is expressed as
  \begin{equation}
    \| \vec{ds} \|_R^2 = \mu_1 x_1^2 + \mu_2 x_2^2 \mbox{ .}
    \label{eqn:vector-length-eigenvalues}
  \end{equation}
  This shows that changing the Euclidean metric with a Riemannian
  metric locally expands or contracts the space along $\vec{v}_1$ and
  $\vec{v}_2$ with ratios $1/\sqrt{\mu_1}$ and
  $1/\sqrt{\mu_2}$. Suppose now that $L_E$ replaced by $L_R$ in
  (\ref{eqn:constraints}). In a place where the edges of the model are
  aligned with $\vec{v}_1$ this yields
  \begin{equation}
    \frac{\delta}{\sqrt{\mu_1}} \leq L_E(e) \leq \frac{\zeta \delta}{\sqrt{\mu_1}}
    \mbox{ .}
    \label{eqn:edge-length-eigenvalues}
  \end{equation}
  Of course a similar inequality holds in the orthogonal
  direction. This shows that (from a Euclidean point of view) the
  vertex density on the mesh of the model is increased by a ratio
  $\sqrt{\mu_1}$ in the direction of $\vec{v}_1$ and by a ratio
  $\sqrt{\mu_2}$ in the direction of $\vec{v}_2$.  Therefore, at a
  given point of the image space, a direct control over the vertex
  density on the deformable mesh is obtained by properly tweaking
  $\vec{v}_1$, $\vec{v}_2$, $\mu_1$ and $\mu_2$ in accordance with
  underlying image features.

  Although these eigenvectors and eigenvalues could be given by a
  user, it is a tedious and complicated task. It is therefore much
  more attractive and efficient to have them selected automatically.
  The subsequent paragraphs discuss this problem and describe a method
  to build the metric directly from the input image in such a way that
  the vertex density of the mesh adapts to the geometry of image
  components and no longer depends on the resolution of input data.

  This property is interesting because the model complexity is made
  independent from the image resolution and is defined instead only by
  the geometrical complexity of the object to recover. Now, the
  geometrical complexity of an object embedded in an image cannot
  exceed the image resolution. Furthermore, since objects do not have
  high curvatures everywhere on their boundary, this complexity is
  generally much smaller.

  \subsection{Definition of metrics}
  Two cases have to be considered:
  \begin{enumerate}
  \item the case for which the model has converged, and for which we
    expect it to follow image contours,
    \label{case-1}
  \item and the case for which the model is still evolving. Thus parts
    of the curve may be far away of image details or cross over
    significant contours.
    \label{case-2}
  \end{enumerate}

  In case \ref{case-1}, the length of its edges is determined by (i)
  the geometrical properties of the recovered image components and
  (ii) the certainty level of the model position. More precisely, the
  length of edges is an increasing function of both the strength and
  curvature of the underlying contours.

  In case \ref{case-2}, two additional sub-cases are possible. 
  \begin{itemize}
  \item If the model crosses over the boundary of image
    components, the vertex density on the curve is increased. By this
    way the model is given more degrees of freedom to get aligned with
    the contour. 
  \item In a place with no image feature ({\it i.e.\/} far away from
    image contours) vertex density is kept as low as possible. As a
    consequence, the number of vertices and hence the computational
    complexity decreases. Moreover since edge length is increased,
    vertices are allowed to travel faster. The number of iterations
    required to reach the rest position of the model is thus reduced.
  \end{itemize}

\begin{figure}[p]
    \begin{center}
      \epsfig{width=0.98\textwidth, file=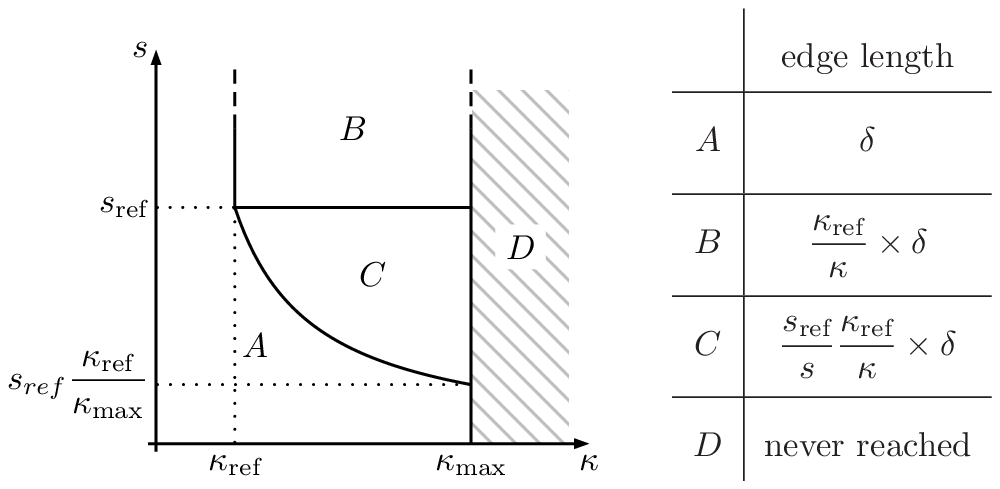}
    \end{center}
    
    \caption{Edge length (up to a factor at most $\zeta$) depending on
      the strength $s$ and curvature $\kappa$ of the underlying
      contour. It is assumed that edges run along the contour. In
      region $A$ contours are too weak or too straight. Therefore,
      edges keep their maximum length. In region $B$ contours are
      considered as reliable and have a curvature higher than the
      reference curvature $\kappa_{\text{ref}}$. The length of the
      edges increases linearly with the radius of curvature of
      underlying contours. In region $C$ contours have a varying
      reliability and have a curvature higher than the reference
      curvature. The length of edges depends on both $s$ and
      $\kappa$. The separation between regions $A$ and $C$ corresponds
      to contours for which
      $\frac{\kappa}{\kappa_{\text{ref}}}\frac{s}{s_{\text{ref}}}=1$. It
      corresponds to contours for which curvature and/or strength fall
      too low to let the model increase its vertex density safely.
    }

    \label{fig:contour-configurations}
  \end{figure}

  To obtain these properties, the eigenstructure of the metric is
  chosen as follows
  \begin{equation}
    \left\{
    \begin{array}{lcl}
      \vec{v}_1 = \vec{n}      & \; \mbox{and} \; & 
      \mu_1=\left[ 
	\displaystyle \frac{s^2}{{s_{\text{ref}}}^2} \times \frac{{\kappa_{\text{max}}}^2}{{\kappa_{\text{ref}}}^2} 
	\right]_{1, \frac{{\kappa_{\text{max}}}^2}{{\kappa_{\text{ref}}}^2}} \\
      \vec{v}_2 = \vec{n}^\bot & \; \mbox{and} \; & 
      \mu_2=\left[ 
	\displaystyle \frac{\kappa^2}{{\kappa_{\text{max}}}^2} \times \mu_1 
	\right]_{1,{\mu_1}}
    \end{array}
    \right. \mbox{ ,}
    \label{eqn:metric-definition}
  \end{equation}
  where~$\vec{n}$ denotes a vector normal to the image contour, and
  the notation~$[\cdot]_{a,b}$ constrains its arguments between the
  bounds~$a$ and~$b$. The parameters~$s$ and~$\kappa$ respectively
  denote the strength and the curvature of contours at the
  investigated point of the image. The parameter $\kappa_{\text{max}}$
  corresponds to the maximum curvature that is detected in the input
  image. The different possible situations are depicted on
  Fig.~\ref{fig:contour-configurations}. Computing these parameters
  directly from images is not straightforward and is discussed in
  Sect.~\ref{sect:computing-strength-curvature}.

\begin{figure}

    \begin{center}
      \epsfig{width=0.98\textwidth, file=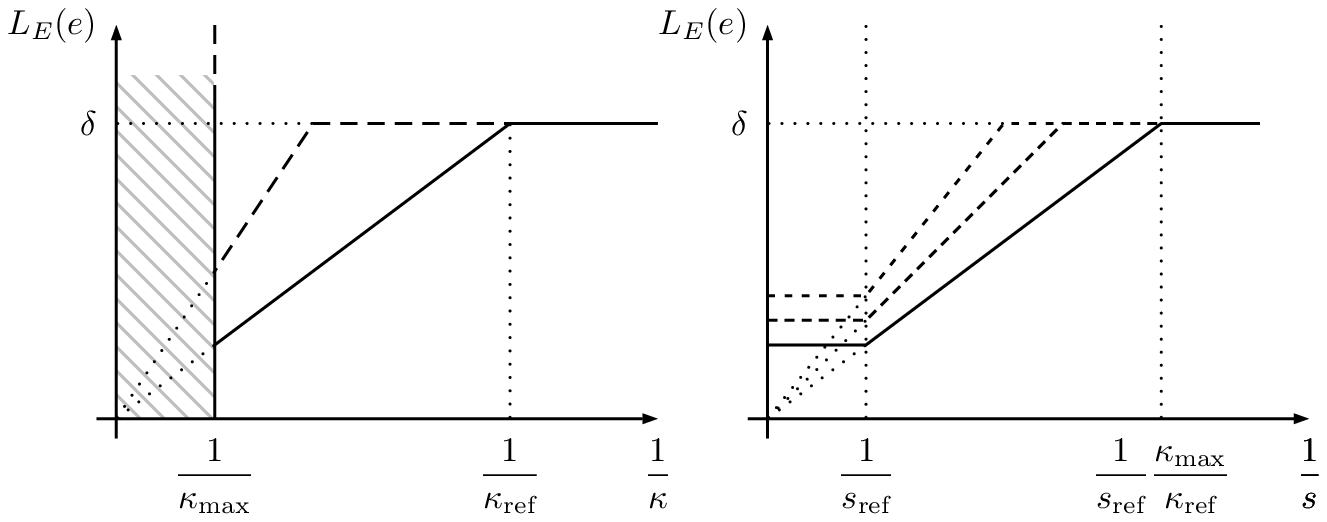}
    \end{center}
    
    \caption{ Left: edge length (up to a $\zeta$ factor) as a function
      of the radius of curvature of the underlying contour. The solid
      line corresponds to a reliable contour ($s \geq
      s_{\text{ref}}$), the dashed line corresponds to a weaker
      contour ($s \leq s_{\text{ref}}$). The hatched part of the graph
      cannot be reached since estimated curvatures cannot exceed
      $\kappa_{\text{max}}$.  
      Right: edge length (up to a factor at most $\zeta$) as a
      function of the strength of the underlying contour. The solid
      line corresponds to a contour with the highest possible
      curvature ($\kappa=\kappa_{\text{max}}$). The dashed lines corresponds
      to less curved contours ($\kappa \leq \kappa_{\text{max}}$).  
      In both figures, it is assumed that edges run along the contour.
      }
    
    \label{fig:edge-length-curves}
  \end{figure}

  The parameter~$s_{\text{ref}}$ is user-given. It specifies the
  strength over which a contour is assumed to be reliable. If an edge
  runs along a reliable contour, then its length is determined only by
  the curvature of the contour (see region~$B$ in
  Fig.~\ref{fig:contour-configurations} and
  Fig.~\ref{fig:edge-length-curves}-left). In other cases the length
  of edges decreases as the contour gets weaker.

  The parameter $\kappa_{\text{ref}}$ is a reference curvature for
  which the length of edges is allowed to vary between $\delta$ and
  $\zeta \delta$ only. Below this curvature contours are assumed to be
  straight and the length of edges remains bounded between $\delta$
  and $\zeta \delta$. Along more curved contours, the length of edges
  increases linearly with the radius of curvature (see
  Fig.~\ref{fig:edge-length-curves}-left). 
  This parameter is easily
  computed from the minimal length allowed by the user for the edges
  (see Fig.~\ref{fig:contour-configurations}):
  \begin{equation}
    \frac{\kappa_{\text{ref}}}{\kappa_{\text{max}}} \delta = l_{\text{min}}
    \mbox{,}
  \end{equation}
  where $l_{\text{min}}$ denotes the chosen minimal length. To take
  advantage of all the details available in the image, it is usual to
  set $l_{\text{min}}$ to the half width of a pixel.

  Note that all the parameters are squared to compensate for the
  square root in (\ref{eqn:edge-length-eigenvalues}). By this way the
  length of edges varies linearly with both $1/s$ and $1/k$.

  \subsection{Influence of the resolution of input images}
  For input images with sufficiently high sampling rates, both $s$ and
  $\kappa$ are determined only by the geometry of image
  components. Consequently, the vertex density during the evolution of
  the model and after convergence are completely independent from the
  image resolution (see experimental results on
  Fig.~\ref{fig:curves}).

  If the sampling rate is too low to preserve all the frequencies of
  objects, contours are smoothed and fine details may be damaged. As a
  result, $s$ is underestimated over the whole image and $\kappa$
  along highly curved parts of objects. In these areas, the
  insufficient sampling rate induces longer edges and details of
  objects cannot be represented accurately. However, these fine
  structures are not represented in input images. As a consequence, it
  is not worth increasing the vertex density since small features
  cannot be recovered even with shorter edges.

  In featureless regions, or along straight object boundaries, the
  length of the edges depend neither on $s$ nor on $\kappa$ and
  remains bounded between $\delta$ and $\zeta \delta$ (see
  Fig.~\ref{fig:contour-configurations}). As a result the improper
  sampling rate of the image has no impact on the vertex density on
  the deformable curve in these regions, which remains coarse.

  As a result of this behavior, the segmentation process is able to
  take advantage of all finest details that can be recovered in the
  image. Indeed, when the resolution of the input image is
  progressively increased, contours and details are restored and $s$
  and $\kappa$ get back their actual value. As a consequence the
  lengths of edges progressively decrease in image parts with fine
  details while remaining unchanged elsewhere. At the same time the
  global complexity of the model increases only slightly with the
  resolution of images until all the frequencies of image components
  are recovered. If the image is oversampled, $s$ and $\kappa$ are
  left invariant, and the number of vertices, and hence the complexity
  remains unaffected. 

  These properties are illustrated experimentally on
  Fig.~\ref{fig:curves}, Fig.~\ref{fig:is-examples}
  and~\ref{fig:fractal}.

  \subsection{Computing strength and curvature of contours from images}
  \label{sect:computing-strength-curvature}
  To tailor metrics to enhance and fasten image segmentation we need
  to estimate both the strength $s$ of image contours and their
  curvature $\kappa$.

  Consider a unit vector $\vec{v}$ and $Q_\vec{v}(x,y) = (\vec{v}
  \cdot \gdt{I}(x,y))^2$. This quantity reaches its maximum when
  $\vec{v}$ has the same orientation (modulo $\pi$) as $\vec{\nabla
    I}(x,y)$. The minimum is reached in the orthogonal direction. To
  study the local variations of the image gradient it is convenient to
  consider the average of $Q_\vec{v}$ over a neighborhood. It is
  expressed in a matrix form as
  \begin{equation}
    \overline{Q_\vec{v}}(x,y) = \tsp{\vec{v}} \times
    \overline{\gdt{I} \times \tsp{\gdt{I}}} \times \vec{v} \mbox{ ,}
  \end{equation}
  where $\overline{(\cdot)}$ denotes the average of its argument over
  a neigborhood of point $(x,y)$. The positive-definite symmetrical
  matrix $J = \overline{\gdt{I} \times \tsp{\gdt{I}}}$ is known as the
  {\em gradient structure tensor}. This operator is classically used
  to analyze local structures of images \cite{Kass87b}, since it
  characterizes their local orientations. It is further used for
  texture and image enhancement in anisotropic diffusion schemes
  \cite{Weickert95,Weickert99}.

  Let $\left\{(\vec{w}_1, \xi_1), (\vec{w}_2, \xi_2)\right\}$ denote
  the eigen decomposition of $J$ and assume that $\xi_2 \leq
  \xi_1$. It is easily seen that $\xi_1$ and $\xi_2$ respectively
  correspond to the maximum and minimum values reached by $Q_\vec{v}$
  when the unit vector $\vec{v}$ varies. Eigenvectors indicate the
  directions for which these extrema are reached. Thus, they
  respectively correspond to the average direction of image gradients
  over the investigated neighborhood and to the orthogonal
  direction. The eigenvalues~$\xi_1$ and~$\xi_2$ store information on
  the local coherence of the gradient field in the neighborhood. 
  When $\overline{(\cdot)}$ is implemented as a convolution with a
  Gaussian kernel $g_\rho$ ($\rho$ corresponds the size of the
  investigated neighborhood), 
  the eigenvalues can be combined as follows to build the required
  estimators $s$ and $\kappa$:
  \begin{equation}
    \left\{
    \begin{array}{l}     
      s^2 \simeq \xi_1 + \xi_2 = {\mathrm{Tr}}(J) = g_\rho \ast
      \left( \| \gdt{I} \|^2 \right)\\
      \displaystyle \kappa^2 \simeq \frac{1}{\rho^2} \times \frac{\xi_2}{\xi_1}
    \end{array}
    \right.
    \mbox{ .}  
    \label{eqn:structure-tensor:approximation}
  \end{equation}
  The estimator $s$ is approximately equivalent to the average norm of
  the gradient. The curvature estimator is based on a second order
  Taylor expansion of $I$ along a contour. With this approximation the
  eigenvalues of the structure tensor can be expressed as functions of
  the strength and the curvature of contours. The curvature $\kappa$
  is then easily extracted (see Appendices for more details).

  \section{Experiments}
  \label{sect:experiments}
  
  \subsection{Quality of the curvature estimator}
  \label{sect:estimator-quality}

  This section illustrates the accuracy and robustness of our proposed
  curvature estimator. We investigate the influence of the sizes of
  the Gaussian kernels used to compute the gradient structure tensor
  and we compare our estimator with previous works.

\begin{figure}[p]
    \begin{center}
      \epsfig{width=0.98\textwidth, file=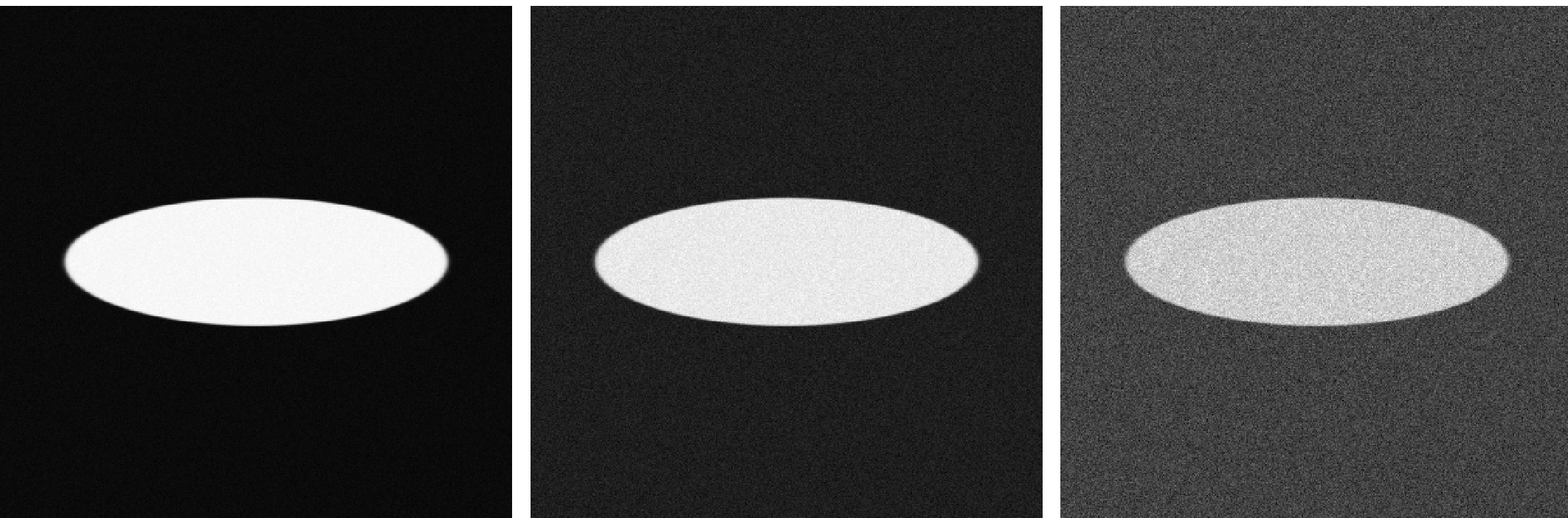}
    \end{center}

    \caption{Images used to test curvature estimators (from left to
      right $PSNR=40\;\mathrm{dB}$, $30\;\mathrm{dB}$ and
      $20\;\mathrm{dB}$).  The curvature is estimated along the border
      of ellipses and are compared with the true curvature for
      different estimators and different values of the
      parameters. Results are presented on
      Fig.~\ref{fig:sigma-influence}-\ref{fig:comparison-1}.  
    }

    \label{fig:ellipses}

  \end{figure}

  For this purpose, we generate images of ellipses with known
  curvature. These images are corrupted with different levels of
  Gaussian noise (see Fig.~\ref{fig:ellipses}). Then curvature is
  computed along ellipses with our estimator and results are compared
  with the true curvature. For a given noise level, the experiment is
  repeated $40$ times. The presented curves show the averages and the
  standard deviations of estimated curvatures over this set of $40$
  test images. Noise levels are expressed using the peak signal to
  noise ratio defined as $PSNR=10 \log \frac{I_{\text{max}}}{\sigma}$
  where $I_{\text{max}}$ is the maximum amplitude of the input signal
  and $\sigma$ is the standard deviation of the noise.

\begin{figure}[p]
    \begin{center}
      \epsfig{width=0.98\textwidth, file=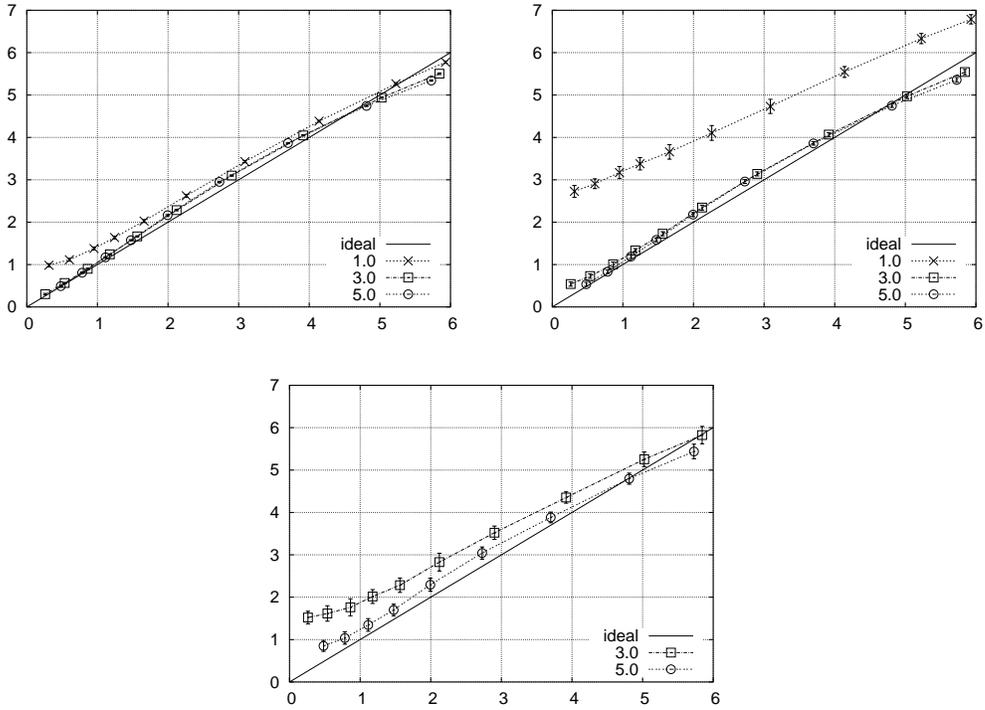}
    \end{center}

    \caption{Estimated curvature as a function of the true curvature
      for different values of $\sigma$ and for $\rho=10$ (the x-axis
      and y-axis respectively correspond to the true and estimated
      curvatures). The three graphics correspond to different noise
      levels: $PSNR=40\;\mathrm{dB}$ (top-left),
      $PSNR=30\;\mathrm{dB}$ (top-right) and $PSNR=20\;\mathrm{dB}$
      (bottom).  }

    \label{fig:sigma-influence}
  \end{figure}

  Fig.~\ref{fig:sigma-influence} illustrates the influence of the
  parameter $\sigma$. As expected, it must be chosen in accordance
  with the noise level in the image. If $\sigma$ is too small, the
  direction of the image gradient changes rapidly in a neighborhood of
  the considered point. As a consequence, the second eigenvalue of the
  structure tensor increases. This explains why curvature is
  overestimated.

\begin{figure}[p]
    \begin{center}
      \epsfig{width=0.98\textwidth, file=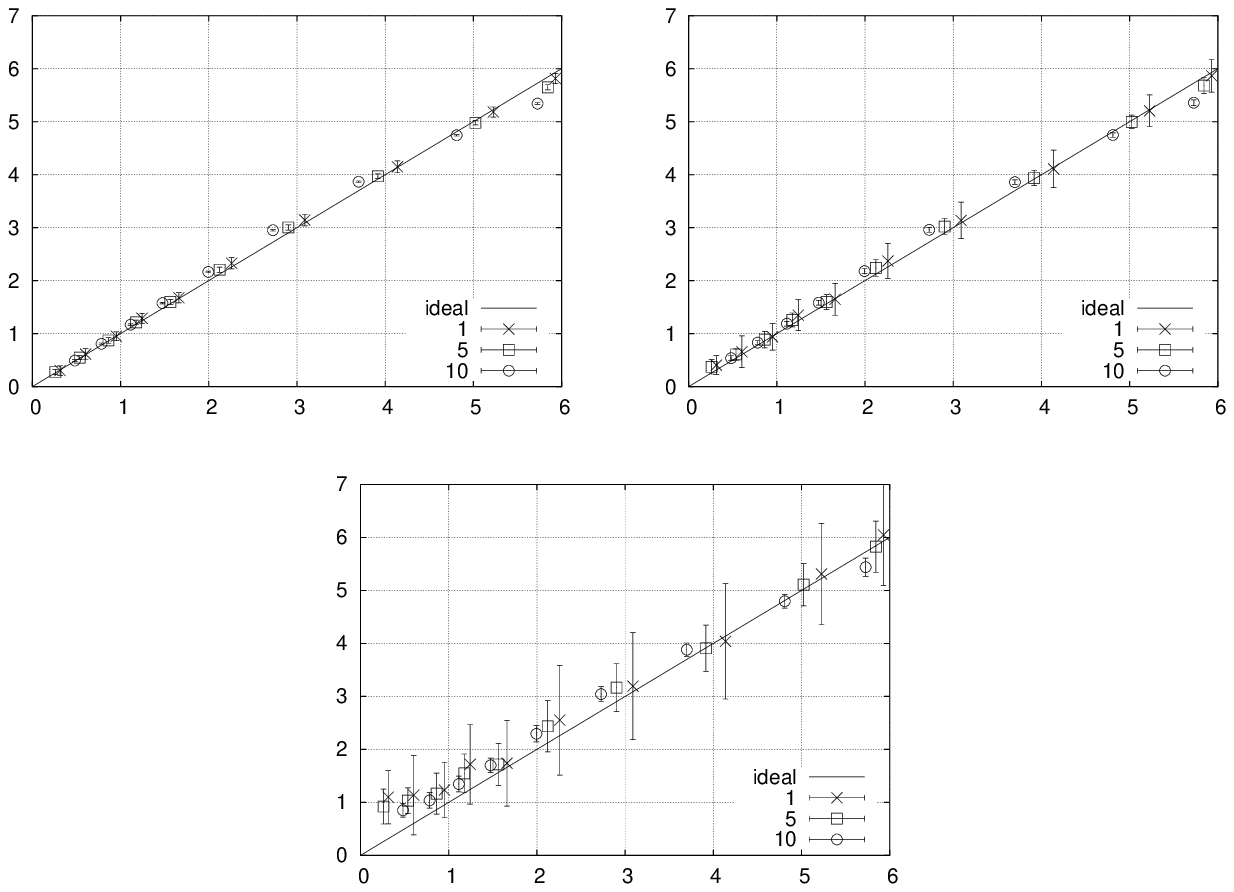}
    \end{center}

    \caption{Estimated curvature as a function of the true curvature
      for $\sigma=5$ and for different values of $\rho$ (abscissa and
      ordinate respectively correspond to the true and estimated
      curvatures). The three graphics correspond to different noise
      levels: $PSNR=40\;\mathrm{dB}$ (top-left),
      $PSNR=30\;\mathrm{dB}$ (top-right) and $PSNR=20\;\mathrm{dB}$
      (bottom).  }

    \label{fig:rho-influence}
  \end{figure}

  The dependency of our estimator on the radius $\rho$ of the local
  integration is depicted on Fig.~\ref{fig:rho-influence}. The
  presented curves show that this parameter has an influence only for
  images with strong noise. Indeed, contour information has to be
  integrated over much larger neighborhoods to mitigate the influence
  of noise.

  In addition, our estimator is compared with two methods which both
  involve the computation of the second derivatives of the input
  image:
  \begin{itemize}
  \item the naive operator which simply computes the curvature of
    isocontours of the image as
    \begin{equation}
      \kappa = \mathrm{div}\left( \frac{\gdt{I}}{\|\gdt{I}\|} \right) =
      \frac{I_{xx} {I_y}^{2} - 2 I_{xy} I_x I_y + I_{yy}
	{I_{x}}^2}{\left( {I_x}^2 + {I_y}^2\right)^{\frac{3}{2}}} \mbox{
	,}
      \label{eqn:curvature-naive}
    \end{equation}
    
  \item the more elaborate estimator proposed by Rieger \textit{et
    al.} \cite{Rieger02}. 
  \end{itemize}
  The latter method consists in computing the derivative of the
  contour orientation in the direction of the contour. Contour
  orientation is computed (modulo~$\pi$) as the
  eigenvector~$\vec{w}_1$ of the gradient structure tensor (which
  corresponds to the largest eigenvalue). Since orientation is only
  known modulo~$\pi$, the vector~$\vec{w}_1$ is converted into an
  appropriate continuous representation (using Knutsson mapping) prior
  to differentiation.
  
\begin{figure}[p]
    \begin{center}
      \epsfig{width=0.98\textwidth, file=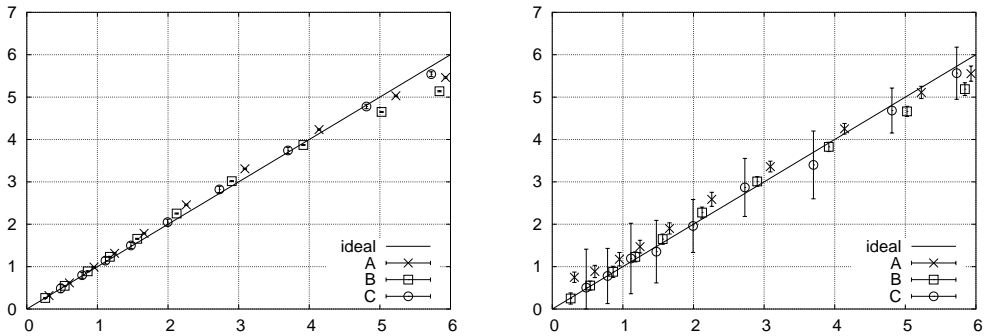}
    \end{center}
    
    \caption{ Comparison of curvature estimators along the contour of
      a noisy ellipse (left $PSNR=40\;\mathrm{dB}$, right
      $PSNR=20\;\mathrm{dB}$). (A) our estimator, (B) Rieger's
      estimator, (C) naive estimator.}

    \label{fig:comparison-1}
  \end{figure}

  As shown on Fig.~\ref{fig:comparison-1} all these estimators provide
  fairly equivalent results along a contour. Note however that the
  naive estimator is much more sensitive to noise than the
  others. 

\begin{figure}[p]
    \begin{center}
      \epsfig{width=0.98\textwidth, file=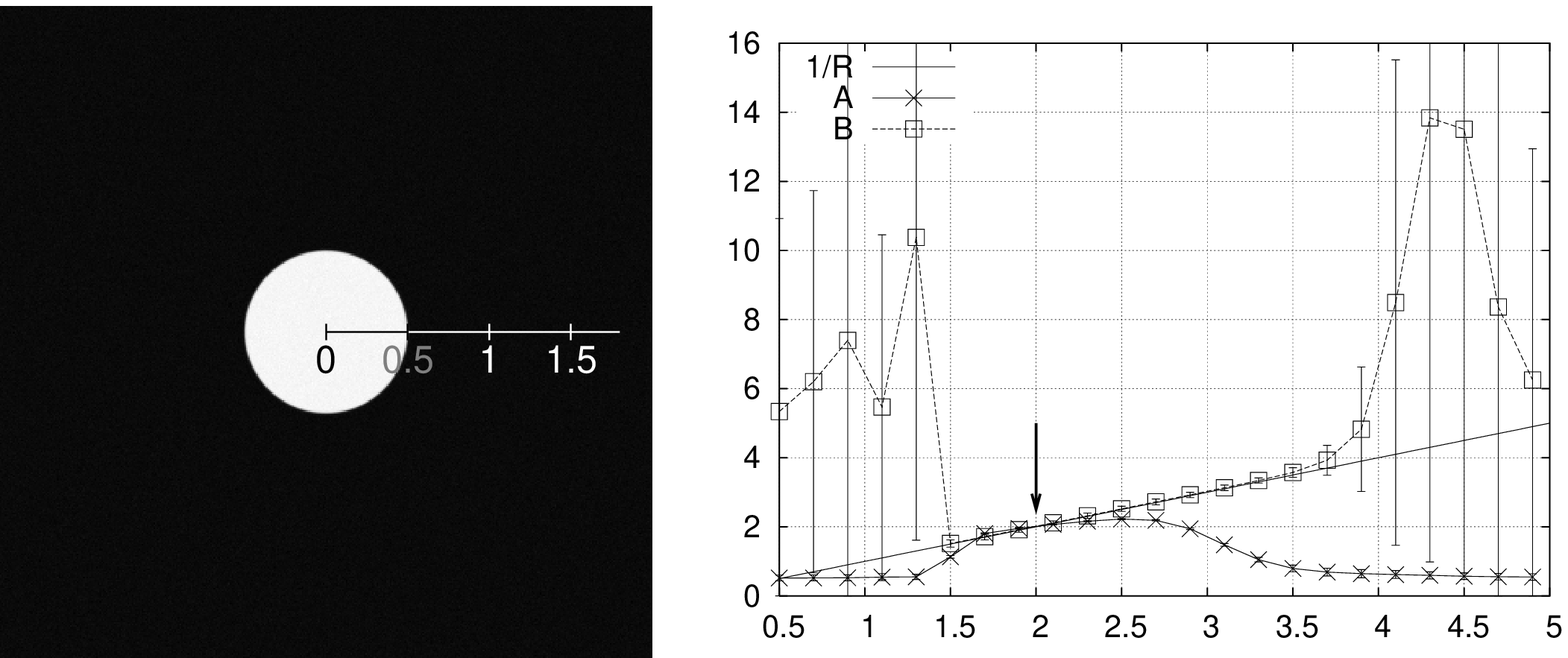}
    \end{center}
    
    \caption{ Comparison of curvature estimators.  Left: test image
      ($PSNR=40\;\mathrm{dB}$). Right: curvature estimated along the
      radius drawn on the left figure. (A) our estimator, (B) Rieger's
      estimator. The solid line represents the inverse of the distance
      to the center of the circle and the arrow indicates the position
      of the contour. The naive operator is not displayed since it is
      too unstable.  }
    
    \label{fig:comparison-2}
  \end{figure}

  These estimators were also tested in places without image
  features. As depicted on Fig.~\ref{fig:comparison-2}, both the naive
  estimator and the one of Rieger {\it et al.\/} become unstable. The
  naive estimator fails because the denominator in
  (\ref{eqn:curvature-naive}) falls to zero and because second
  derivatives are very sensitive to noise. Rieger's method can neither
  be used. Indeed, in a region without significant contour, the
  eigenvector~$\vec{w}_1$ of the gradient structure tensor is only
  determined by noise and thus exhibits rapid variations. Computing
  its derivatives results in a spurious evaluation of the
  curvature. This justifies the use of our estimator, which, in
  addition, requires less computations than Rieger's method since it
  estimates curvature directly from the eigendecomposition of the
  gradient structure tensor and does not involve their derivatives.

  \subsection{Parameter selection for a new image segmentation}
  \label{sect:parameter-selection}

  Given a new image, we have to adjust some parameters to exploit at
  best the potentialies of the proposed approach. We follow the steps
  below:

  \begin{enumerate}

    \item The image structure tensor is computed: it provides the
    contour intensities $s$, the curvatures $k$ and the local metrics;
    the maximal curvature $k_\text{max}$ as well as the maximal
    intensity $s_\text{max}$ follow immediately.

    \item The user chooses the minimal and the maximal edge
    lengths $l_\text{min}$ and $l_\text{max}$ for the
    model. Typically, the length $l_\text{min}$ is half the size of a
    pixel (a better precision has no sense given the input data) and
    the length $l_\text{max}$ is about 50 times $l_\text{min}$ for
    real data. 

    \item The user then selects the reference contour intensity
    $s_\text{ref}$ which corresponds to reliable contours. A simple
    way is to visualize the thresholding of the image $s$ by the value
    $s_\text{ref}$, and to tune this parameter accordingly. It can
    also be automated for certain images as a percentage of
    $s_\text{max}$ (typically $90\%$).

    \item After that, the procedure is the same as for classical
    snakes: initialisation, selection of energy/force parameters,
    evolution until rest position.

  \end{enumerate}

  From the preceding paragraphs, it is clear that the proposed approach
  does not induce significantly more interaction compared with
  classical snakes.

  \subsection{Behaviour of the deformable model}
  \label{sect:deformable-model-experiments}  

\begin{figure}[p]
    \begin{center}
      \epsfig{width=0.98\textwidth, file=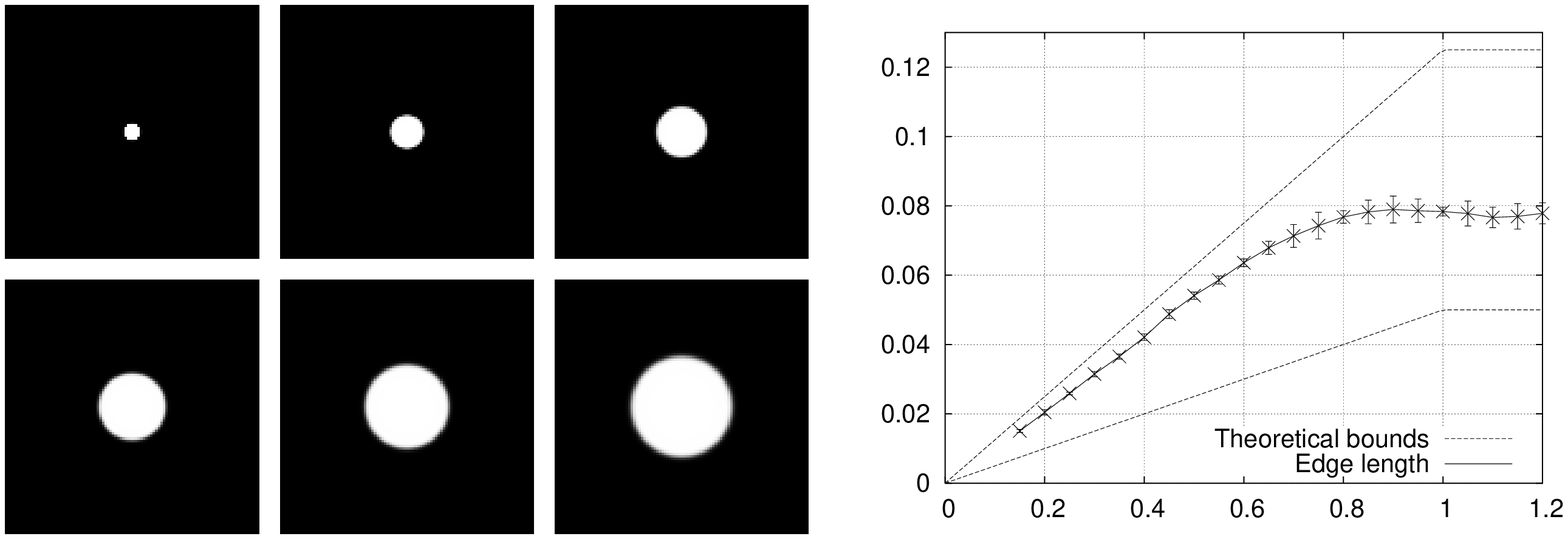}
    \end{center}
    \caption{ Left: test images (circles with known radii). Right:
      edge length as a function of the radius of curvature.
      Dashed lines correspond to the theoretical bounds.  
    }

    \label{fig:circles}
  \end{figure}

  Adaptive vertex density is illustrated in Fig.~\ref{fig:circles}. In
  this experiment images of circles with known radii are generated
  (left part of the figure). For each circle, the image space is
  equipped with a metric which is built as explained in
  Sect.~\ref{sect:building-metrics}. In this example
  $\kappa_{\text{ref}} = 1$ and $s_{\text{ref}}$ is computed from the
  input image as the maximum value of $s$ over the image. Our
  deformable model is then used to segment and reconstruct the
  circles. Once it has converged, the Euclidean lengths of its edges
  are computed. The results are presented on the curve in the right
  part of the figure. They correspond to the expected behavior (see
  Fig.~\ref{fig:edge-length-curves}).

\begin{figure}[p]
    \epsfig{width=0.98\textwidth, file=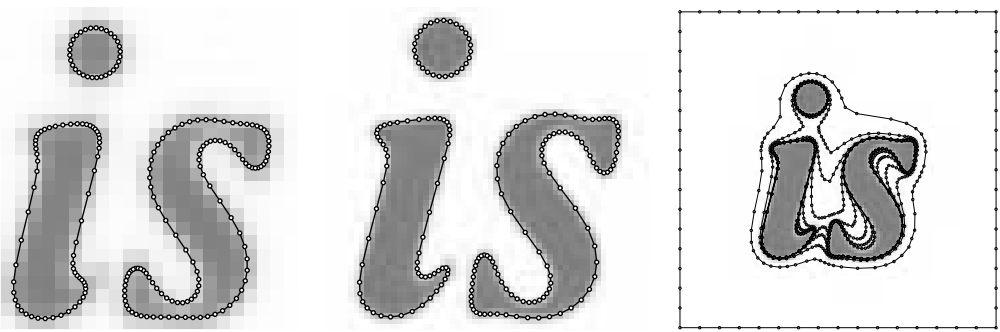}
    \caption{ Left, center: reconstruction of identical objects
      sampled at resolutions 40$\times$40 and 100$\times$100. Right:
      evolution of the deformable model every 50 iterations. The outer
      square corresponds to the initial position of the model.
    }
    \label{fig:is-examples}
  \end{figure}

\begin{figure}
    \begin{center}
      \epsfig{width=0.98\textwidth, file=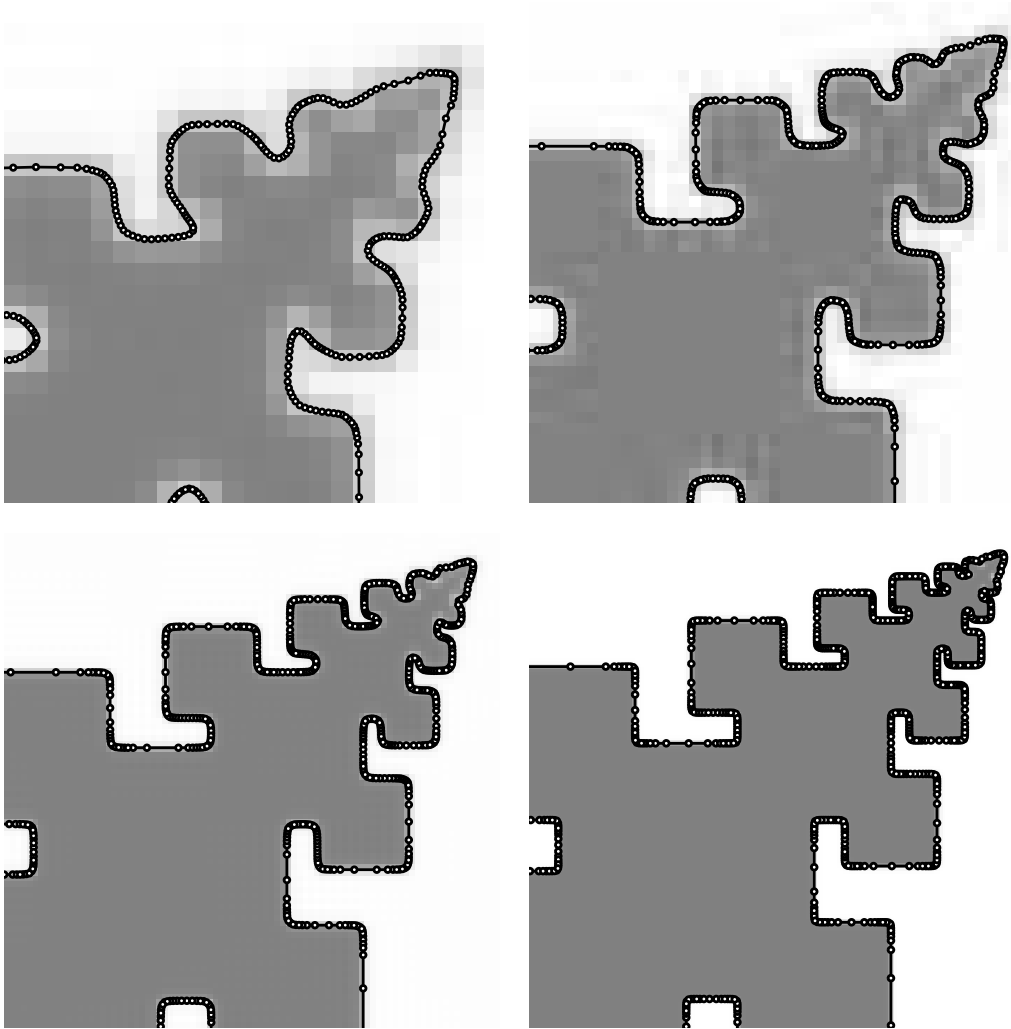}
    \end{center}

    \caption{ Segmentation/Reconstruction of the same object sampled
      at increasing resolutions (50$\times$50, 100$\times$100,
      200$\times$200 and 400$\times$400). For the four images all the
      parameters used to build the metric or attract the model toward
      object boundaries are identical. Please note that the deformable
      model automatically adapts its resolution to represent available
      image features as well as possible, while optimizing the number
      of vertices.
    }

    \label{fig:fractal}
  \end{figure}

\begin{figure}[p]
    \begin{center}
      \epsfig{width=0.98\textwidth, file=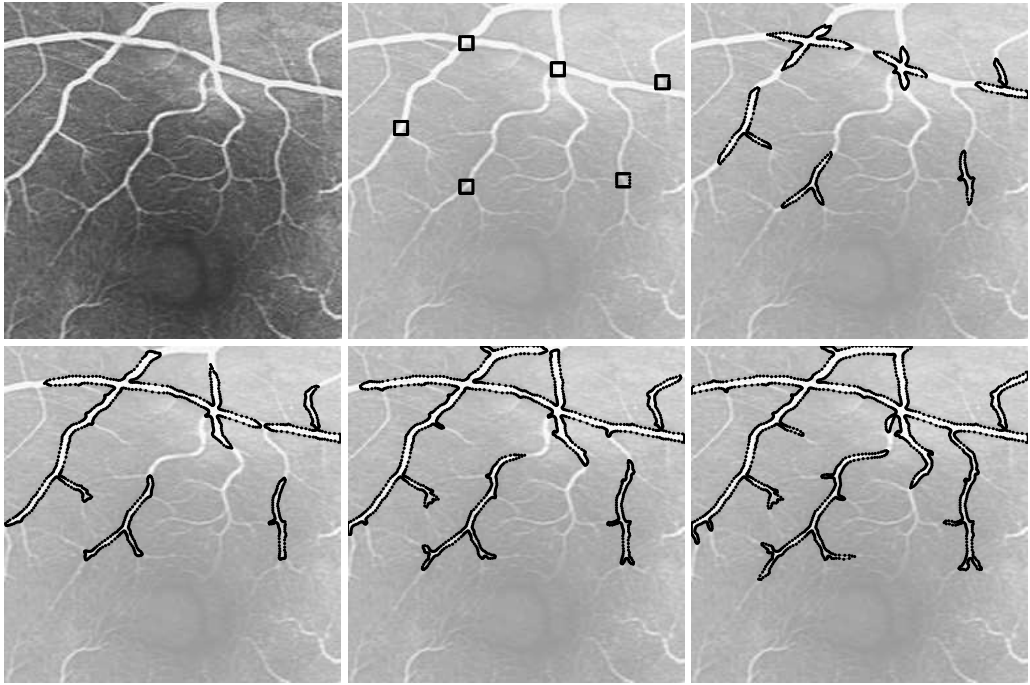}
    \end{center}

    \caption { Segmentation process on an angiography.  Top-left:
      input image. Other images: steps of the evolution of the
      deformable curve. The model is driven by an inflation force
      which stops when the local gray level is lower that the average
      gray-level in a neighborhood. Please note the topology changes
      when parts of the deformable curve collide. Computation times
      are given on Table~\ref{tab:computation-times} }
    \label{fig:angio-iterations}
  \end{figure}

\begin{figure}[p]    
    \begin{center}
      \epsfig{width=0.98\textwidth, file=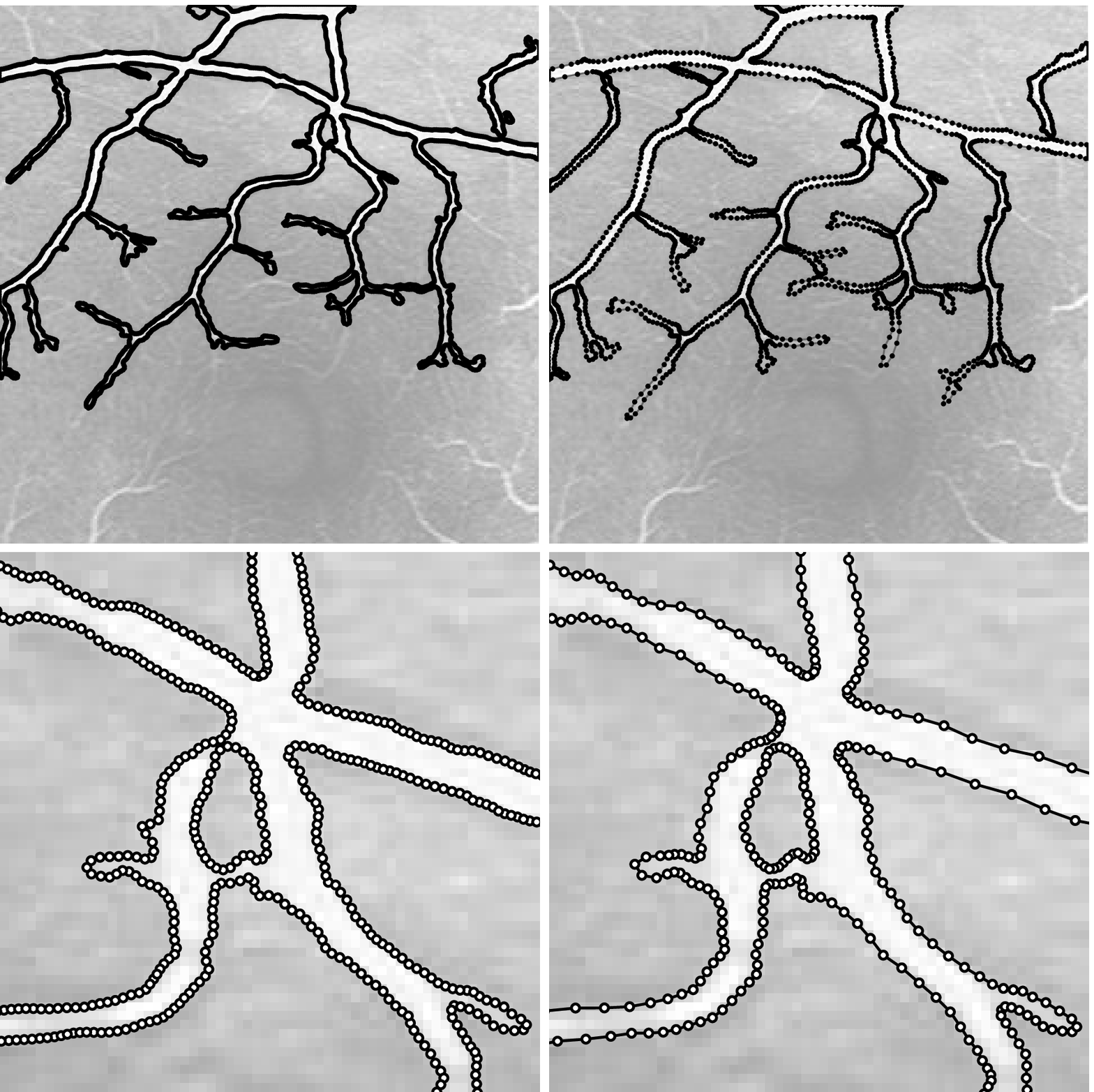}
    \end{center}
    
    \caption { Segmentation of the angiography shown on
      Fig.~\ref{fig:angio-iterations}.  Left: results without
      adaptation. Right: results with a metric built as described in
      \ref{sect:building-metrics}.  Top: final result. Bottom:
      detailed view in a region which exhibits much adaptation of the
      vertex density as well as a complex topology. Please note how
      the length of edges is adapted according to the structures found
      in the input image and how this enables the deformable model to
      enter small gaps and recover structures finer than its edges.
      Computation times are given in Table~\ref{tab:computation-times}.
    }
    \label{fig:angio-example}
  \end{figure}

  Adaptive vertex density is also visible in
  Fig.~\ref{fig:is-examples}-\ref{fig:angio-example}. As expected,
  changing the metric increases vertex density along highly curved
  parts of image components. As a result, the description of the shape
  of objects is enhanced while the number of vertices is optimized.

\begin{figure}
    \begin{center}
      \epsfig{width=0.48\columnwidth, file=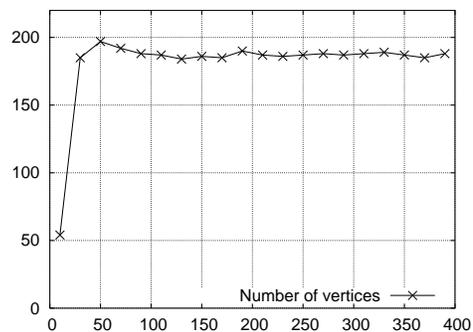}
    \end{center}
    
    \caption{ Final number of vertices on the curve depending on the
      resolution of input image. The segmentation/reconstruction results
      as well as the evolution of the model are shown on
      Fig.~\ref{fig:is-examples}.  
    }
    \label{fig:curves}
  \end{figure}

  Independence with respect to the resolution of input images is shown
  on Fig.~\ref{fig:curves}. Our model was tested on images of objects
  sampled at different rates (see Fig.~\ref{fig:is-examples}). As
  expected, the number of vertices is kept independent from the
  resolution of the input image, as far as the sampling rate ensures a
  proper representation of the highest frequencies present in the
  signal. If this condition is not satisfied, as on
  Fig.~\ref{fig:fractal}, the model uses only the available
  information. If the resolution is increased, the length of the edges
  of the model remain unchanged, except in parts where the finer
  sampling rate of the image allows to recover finer features.

  Fig.~\ref{fig:is-examples} and Fig.~\ref{fig:angio-iterations}
  demonstrate the ability of our model to dynamically and
  automatically adapt its topology. Note that the proposed way to
  build the metric is especially well suited to objects with thin and
  elongated features.  With previous approaches
  \cite{Delingette00,McInerney95} automated topology changes can only
  be achieved using grids with a uniform resolution determined by the
  thinest part of objects. Their complexities are thus determined by
  the size of the smallest features to be reconstructed. The involved
  computational effort is therefore wasteful since much more vertices
  are used than required for the accurate description of objects. In
  contrast, replacing the Euclidean metric with a metric designed as
  described in this paper virtually broaden thin structures. As a
  consequence, even for large values of $\delta$, the inequality
  (\ref{eqn:collision-detection}) where $d_E$ has been replaced by
  $d_R$ is not satisfied for two vertices $u$ and $v$ located on
  opposite sides of long-limbed parts of image
  components. Self-collisions are thus detected only where they really
  occur. At the same time, the number of vertices is kept independent
  from the size of the finest details to be recovered.

  Fig.~\ref{fig:angio-iterations}-\ref{fig:angio-example} illustrate
  the behavior of our deformable model on biomedical images. The input
  image (Fig.~\ref{fig:angio-iterations} top-right) is a fluorescein
  angiogram that reveals the structure of the circulatory system of
  the back of an eye. In addition to the classical regularizing
  forces, the vertices of the active contour undergo an
  application-specific force designed to help recovering blood
  vessels. This force pushes vertices in the direction of the outer
  normal and stops when the local gray level of the image falls under
  the average gray level over a neighborhood. More formally, the force
  $\vec{F}_v$ undergone by a vertex $v$ is defined as
  \begin{equation}
    \vec{F}_v = \left( I(v) - (g_\tau \ast I)(v) \right) \times \vec{n}_v \mbox{ .}
  \end{equation}
  where $I$ is the input image, $g_\tau$ is a Gaussian filter used to
  estimate the average gray level over a neighborhood, and $\vec{n}_v$
  is the outer normal to the deformable curve at vertex $v$. The
  presented results demonstrate the possibility to use additional
  forces designed to extend or improve deformable models
  \cite{Xu98,Cohen91,Yu02}. Furthermore, computation times are given
  in Table~\ref{tab:computation-times}. They show that reducing the
  number of required iterations and the number of vertices largely
  compensate for the time used to compute of the metric. These
  computation times are given in
  Table~\ref{tab:metric-computation-time} for different size of input
  images.

  At last, since the space is expanded only in the vicinity of image
  contours, vertices travel faster in parts of the image without
  feature. When approaching object boundaries, the deformable curve
  propagates slower and progressively refines. The advantage is
  twofold. First, the number of iterations required for the model to
  reach its rest position is reduced. Second, the cost of an iteration
  is reduced for parts of the deformable curve far away from image
  features, namely parts with a low vertex density. By this way a lot
  of computational complexity is saved when the deformable model is
  poorly initialized. This is especially visible on
  Fig.~\ref{fig:is-examples} (right) where the position of the model
  has been drawn every 50 iterations.

  \begin{table}[tb]

    \caption{ Number of iterations and time required to reach
      convergence. The table also indicates how many vertices are used
      to represent the whole shapes shown in
      Fig.~\ref{fig:angio-example}. The two last columns describe the
      minimum and maximum length (in pixels) of an edge of the
      deformable model.  
      \label{tab:computation-times}
    }

    \begin{center}
      \begin{tabular}{l|c|c|c|c|c}
	& iterations & total time (s) & vertices & 
	\parbox{0.15\textwidth}{
	  \centering min. edge length} & 
	\parbox{0.15\textwidth}{
	  \centering max. edge length}  \\
	\hline
	uniform  & 350 & 78.5 & 3656 & 0.5 & 1.25\\
	adaptive & 250 & 37.35 (+1.24) & 2065 & 0.35 & 25
      \end{tabular}
    \end{center}

  \end{table}

  \begin{table}[tb]
    \caption{Computation times required to build the metric for
      different sizes of input images. 
      \label{tab:metric-computation-time}
    }

    \begin{center}
      \begin{tabular}{l|ccccccc}
	resolution of input image &  100 &  150 &  200 & 250  & 300  & 350  &  400\\
	\hline
	computation time (s)      & 0.16 & 0.36 & 0.65 & 1.00 & 1.45 & 1.98 & 2.58
      \end{tabular}
    \end{center}
  \end{table}

  \section{Conclusion}
  \label{sect:conclusion}

  We presented a deformable model that adapts its resolution according
  to the geometrical complexity of image features.  It is therefore
  able to recover finest details in images with a complexity almost
  independent from the size of input data. Admittedly, a preprocessing
  step is required to build the metric. However, involved
  computational costs are negligible and, as a byproduct, these
  precomputations provide a robust gradient estimator which can be
  used as a potential field for the deformable model. Most of the
  material used in our presented deformable model has a
  straightforward extension to higher dimensions \cite{Lachaud03,Taton04}. We
  are currently working on the extension of the whole approach to 3D
  images.

  \appendix

  \section{Second order approximation of contours}
  \label{sect:2nd-order-approximation}
  
  In this section, we consider a contour that is tangent to the $x$
  axis at the origin. This is expressed as $I_x(0,0) = 0$ and
  $I_y(0,0)=s$, where $I_x$, $I_y$ and $s$ denote the partial
  derivatives of $I$ and the strength of the contour.

  From the definition of a contour as a maximum of the norm of the
  gradient in the gradient direction follows $ \left. \frac{\partial
    \left( \| \vec{\nabla I} \| \right)}{\partial \vec{\nabla I}}
  \right|_{(0, 0)} = 0 $. Once expanded, this leads to
  $I_{yy}(0,0)=0$.

  Let $\vec{t}$ and $\vec{n}$ denote the vectors tangent and normal to
  the investigated contour: $\vec{n} = \frac{\gdt{I}}{\|\gdt I\|}$ and
  $\vec{t} = \vec{n}^{\bot}$. From the definition of curvature follows
  $\pder{\vec{n}}{\vec{t}} = \kappa \vec{t}$. Replacing $\vec{t}$ and
  $\vec{n}$ by their expression as functions of $I$, and then
  expanding and evaluating this expression at point $(0, 0)$ yields
  $I_{xx}(0, 0) = s \kappa$.

  From the above statements we get a second order Taylor expansion of
  $I$ as
  \begin{equation}
    I(x,y) - I(0,0) =
    s \left( y + \frac{1}{2} \kappa x^2 \right) + I_{xy}xy + o(x^2, y^2) \mbox{ .}
    \label{eqn:parabolic-approximation-varying-s}
  \end{equation}
    
  In addition, if we assume that the strength of the contour remains
  constant along the contour, we get $\left. \frac{\partial \left( \|
    \vec{\nabla I} \|\right)}{\partial \vec{\nabla I}^{\bot}}
  \right|_{(0,0)} = 0$. Expanding the previous expression leads to
  $I_{xy}(0,0)=0$.
  
  With this additional hypothesis, $I$ may be rewritten as
  \begin{equation}
    I(x,y) - I(0,0) =
    s \left( y + \frac{1}{2} \kappa x^2 \right) + o(x^2, y^2) \mbox{ .}
    \label{eqn:parabolic-approximation}
  \end{equation}

  \section{Structure tensor of a parabolic contour}
  \label{sect:curvature-approximation}
  In this section we compute the eigenvalues of the structure tensor
  along a contour with strength $s$ and with local curvature
  $\kappa$. 

  \subsection{Contour with constant intensity} 
  Following approximation (\ref{eqn:parabolic-approximation}) we
  consider the image $I$ defined as
  \begin{equation}
    I(x,y) = s \left( y + \frac{1}{2} \kappa x^2 \right) \mbox{ .}
    \label{eqn:parabolic-contour}
  \end{equation}
  For symmetry reasons, we know that the eigenvectors of the structure
  tensor $J$ at point $(0, 0)$ are aligned with the $x$ axis and $y$
  axis. In this special case, the eigenvalues of $J$ are given as
  $\xi_1=\overline{{I_y}^2}=\overline{s^2}$ and
  $\xi_2=\overline{{I_x}^2}=\overline{s^2 \kappa^2 x^2}$. If the
  averaging operation over a neighborhood is implemented as a
  convolution with a Gaussian function $g_\rho$, this yields
  $\xi_1=s^2$ and $\xi_2=s^2 \kappa^2 \rho^2$. In practice only
  $\xi_1$, $\xi_2$ and $\rho$ are known. Curvature (up to sign) is
  easily computed from these quantities as
  \begin{equation}
    |\kappa| = \frac{1}{\rho} \sqrt{\frac{\xi_2}{\xi_1}}
    \label{eqn:curvature-estimator}
  \end{equation}

  \subsection{Contour with varying intensity} 
  In this subsection we show that the estimator described in the
  previous paragraph remains valid to estimate the curvature of a
  contour with a varying intensity

  We start with equation
  (\ref{eqn:parabolic-approximation-varying-s})~:
  \begin{equation}
    I(x,y) = s \left( y + \frac{1}{2} \kappa x^2 \right) + I_{xy}xy \mbox{ ,}
  \end{equation}
  from which we get $\xi_1=s^2 + I_{xy}^2$ and $\xi_2 =s^2 \kappa^2
  \rho^2 + I_{xy}\rho^2$.

  The curvature estimation $\hat{\kappa}$ is thus written~:
  \begin{equation}
    \hat{\kappa} 
    = \frac{1}{\rho} \sqrt{\frac{\xi_2}{\xi_1}} 
    = \left(\frac{\kappa^2 s^2 + I_{xy}}{s^2 + I_{xy} \rho^2}\right)^{\frac{1}{2}} \mbox{ .}
  \end{equation}

  If $\kappa = 0$ we get 
  \begin{equation}
    \hat{\kappa} = 
    \left(\frac{1}{\left(\frac{s}{I_{xy}}\right)^2 + \rho^2}\right)^{\frac{1}{2}}
    \mbox{ ,}
    \label{eqn:null-curvature}
  \end{equation}
  Assuming that the contour intensity is
  significantly greater than its linear variation along $x$, we obtain
  $\frac{I_{xy}}{s} \simeq 0$.  
  Replacing in (\ref{eqn:null-curvature}) we get $\hat{\kappa} \simeq
  0$.

  If $\kappa \neq 0$, we get 
  \begin{equation}
    \hat{\kappa} = \kappa +
    \frac{1-\kappa^2\rho^2}{2k} \times \left(\frac{I_{xy}}{s}\right)^2 +
    o\left(\left(\frac{I_{xy}}{s}\right)^3\right) \mbox{.}
  \end{equation}
  As shown before $\frac{I_{xy}}{s} \simeq 0$ for a reliable contour.
  As a consequence, such a contour $\hat{\kappa} \simeq \kappa$, which
  shows that the curvature estimator remains available for contours
  with a varying intensity.

  \section{Implementation issues}
  The gradient structure tensor is implemented as successive
  convolutions of the input image with two Gaussian functions
  $g_\sigma$, $g_\rho$ and their partial derivatives:
  \begin{equation}
    J = g_\rho \ast (\nabla (I \ast g_\sigma) \times \tsp{\nabla (I \ast
      g_\sigma)} ) \mbox{ .}
    \label{eqn:structure-tensor:implementation}
  \end{equation}
  Convolutions are implemented efficiently as a product in the
  frequency domain and could be further improved using recursive
  implementations of Gaussian filters \cite{Deriche92,VanVliet98}.

  The parameter $\sigma$ determines how much the image gets smoothed
  before computing its derivatives. It is thus chosen in accordance
  with the noise level in the image. The parameter $\rho$ determines
  the size of the neighborhood over which the gradient information is
  integrated. The influences of these parameters are studied
  experimentally in Sect.~\ref{sect:estimator-quality}.

  Since the metric has to be computed everywhere in the image our
  estimator must remain stable in regions without contours ({\it
    i.e.\/} in regions where $\xi_1\simeq 0$. Therefore $\kappa$ is
  computed as follows:
  \begin{equation}
    \kappa \simeq \frac{1}{\rho} \sqrt{\frac{\xi_2}{\xi_1+
	\epsilon}}\mbox{ ,}
    \label{eqn:curvature-estimator-final}
  \end{equation}
  where $\epsilon$ is an arbitrary positive constant.  By this way the
  denominator never vanishes and $\kappa$ falls to $0$ in places
  without image structure. In the vicinity of image contours
  $\epsilon$ may be neglected in front of $\xi_1$ and we get back
  estimation (\ref{eqn:curvature-estimator}).  Experimentally, a
  suitable choice for this constant is $\epsilon =
  \frac{1}{10}\xi_1^{\text{max}}$ where $\xi_1^{\text{max}}$ denotes
  the maximum value of $\xi_1$ over the image.


\end{document}